\documentclass[10pt,twocolumn,letterpaper]{article}

\usepackage{cvpr}
\usepackage{times}
\usepackage{epsfig}
\usepackage{graphicx}
\usepackage{amsmath}
\usepackage{amssymb}

\usepackage{booktabs}
\usepackage{tabulary}
\usepackage{makecell}
\usepackage{multirow}
\usepackage{algorithm}
\usepackage{algpseudocode}
\usepackage{enumitem}
\usepackage{array}
\usepackage{float}
\usepackage{wrapfig}
\usepackage{subcaption}

\newcommand{\eat}[1]{}

\usepackage[pagebackref=true,breaklinks=true,letterpaper=true,colorlinks,bookmarks=false]{hyperref}

\cvprfinalcopy 

\def\cvprPaperID{4302} 

\ifcvprfinal\fi
\begin{document}

\title{Learning Instance Occlusion for Panoptic Segmentation}

\author{Justin Lazarow\thanks{ \:indicates equal contribution.} \quad Kwonjoon Lee\footnotemark[1] \quad Kunyu Shi\footnotemark[1] \quad Zhuowen Tu\\
University of California San Diego\\
{\tt\small $\{$jlazarow, kwl042, kshi, ztu$\}$@ucsd.edu}
}

\maketitle
\thispagestyle{empty}

\begin{abstract}
  Panoptic segmentation requires segments of both ``things'' (countable object instances) and  ``stuff'' (uncountable and amorphous regions) within a single output. A common approach involves the fusion of instance segmentation (for ``things'') and semantic segmentation (for ``stuff'') into a non-overlapping placement of segments, and resolves overlaps.
  However, instance ordering with detection confidence do not correlate well with natural occlusion relationship.
  To resolve this issue, we propose a branch that is tasked with modeling how two instance masks should overlap one another as a binary relation. Our method, named OCFusion, is lightweight but particularly effective in the instance fusion process. OCFusion is trained with the ground truth relation derived automatically from the existing dataset annotations. We obtain state-of-the-art results on COCO and show competitive results on the Cityscapes panoptic segmentation benchmark.
\end{abstract}

\vspace{-5mm}
\section{Introduction}
Image understanding has been a long standing problem in both human perception \cite{biederman1987recognition} and computer vision \cite{marr1982vision}. The {\em image parsing} framework \cite{tu2005image} is concerned with the task of decomposing and segmenting an input image into constituents such as objects (text and faces) and generic regions through the integration of image segmentation, object detection, and object recognition. Scene parsing is similar in spirit and consists of both non-parametric \cite{tighe2014scene} and parametric \cite{zhao2016pspnet} approaches.

\begin{figure}[!ht]
\vspace{-1mm}
\begin{center}
\begin{tabular} {c}
\includegraphics[width=0.48\textwidth]{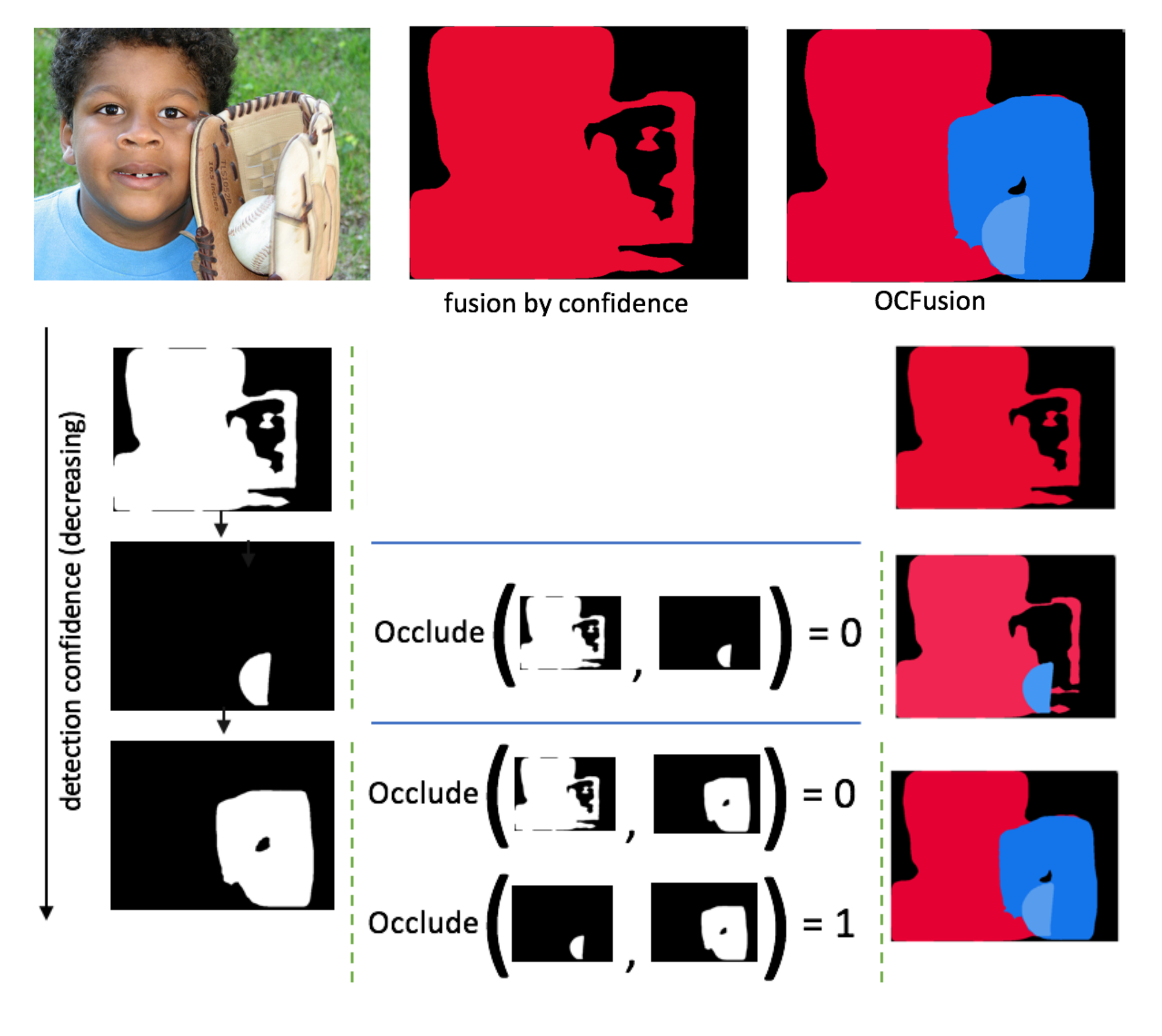}
\end{tabular}
\end{center}
\vspace{-10mm}
\caption{\textbf{An illustration of fusion using masks sorted by detection confidence alone \cite{kirillov2018panoptic} \vs with the ability to query for occlusions (OCFusion; ours).} Occlude$(A,B)=0$ in occlusion head means mask $B$ should be placed on top of mask $A$. Mask R-CNN proposes three instance masks listed with decreasing confidence. The heuristic of \cite{kirillov2018panoptic} 
occludes all subsequent instances after the ``person'', while our method retains them in the final output by querying the occlusion head.}
\vspace{-5mm}
\label{fig:fusion_process}
\end{figure}

\begin{figure*}[!htp]
\vspace{-5mm} 
\begin{center}
\begin{tabular} {c}
\includegraphics[width=1.0\textwidth]{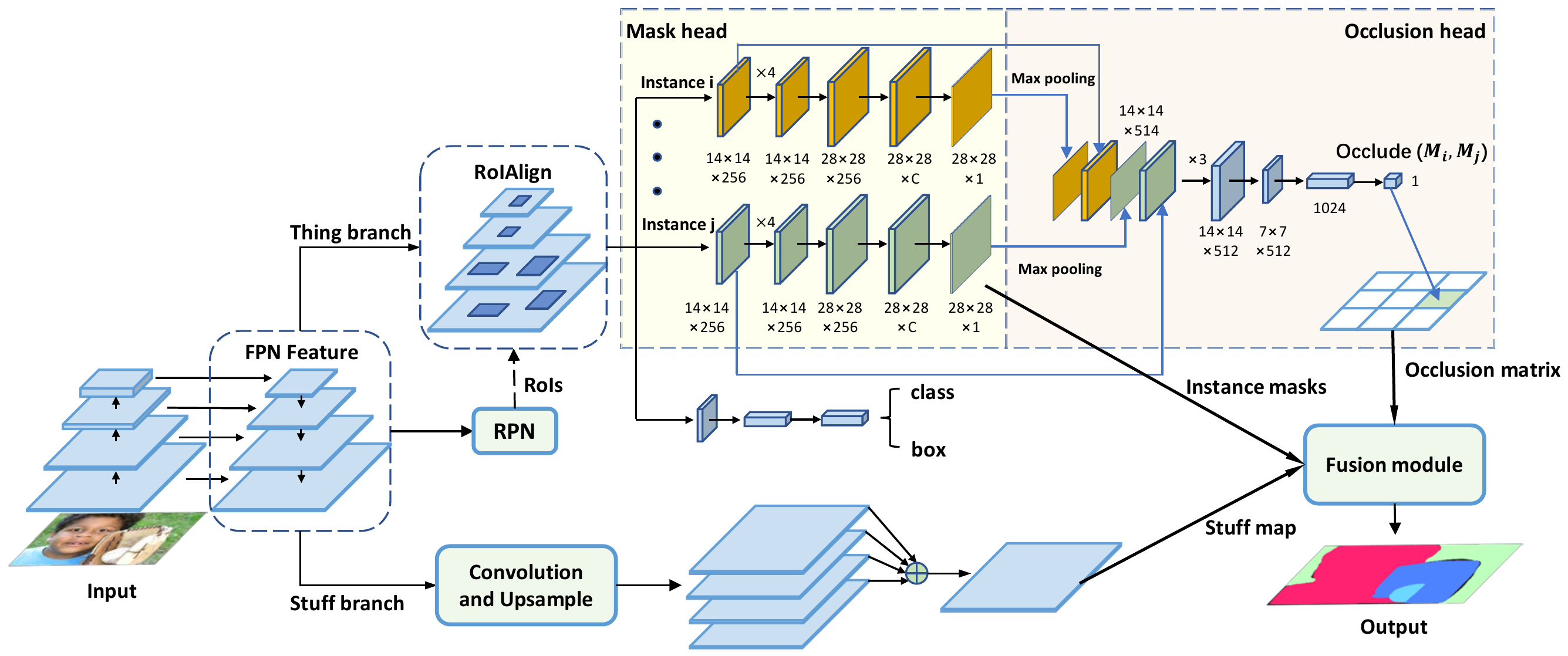}
\end{tabular}
\end{center}
\vspace{-5mm}
\caption{\textbf{Illustration of the overall architecture.} The FPN is used as a shared backbone for both thing and stuff branches. In thing branch, Mask R-CNN will generate instance mask proposals, and the occlusion head will output binary values $Occlude(M_i, M_j)$ (Equation \ref{occlude_equation}) for each pair of mask proposals $M_i$ and $M_j$ with \textit{appreciable} overlap (larger than a threshold) to indicate occlusion relation between them. Occlusion head architecture is described in Section \ref{sec:architecture_of_occlusion_head}. Fusion process is described in \ref{fusion-with_occlusion}.}
\vspace{-4mm}
\label{fig:overall_architecture}
\end{figure*}

After the initial development, the problem of image understanding was studied separately as object detection (or extended to instance segmentation) and semantic segmentation. Instance segmentation \cite{pinheiro2015learning,SharpMask,dai2015instance, liang2018proposal,he2017mask, riemenschneider2012hough,zhang2015instance,jin2016object} requires the detection and segmentation of each \textit{thing} (countable object instance) within an image, while semantic segmentation \cite{shotton2006textonboost, tu2008auto, everingham2010pascal,long2015fully, chen2018deeplab,zheng2015conditional,zhao2016pspnet} provides a dense per-pixel classification without distinction between instances within the same \textit{thing} category. Kirillov \etal \cite{kirillov2018panoptic} proposed the panoptic segmentation task that combines the strength of semantic segmentation and instance segmentation. In this task, \textit{each pixel} in an image is assigned either to a background class (\textit{stuff}) or to a specific foreground object (an \textit{instance} of \textit{things}). 

A common approach for panoptic segmentation has emerged in a number of works \cite{kirillov2019panoptic,li2018attention,xiong2019upsnet} that relies on combining the strong baseline architectures used in semantic segmentation and instance segmentation into either a separate or shared architecture and then \textit{fusing} the results from the semantic segmentation and instance segmentation branches into a single panoptic output. Since there is no expectation of consistency in proposals between semantic and instance segmentation branches, conflicts must be resolved. Furthermore, one must resolve conflicts \textit{within} the instance segmentation branch as it proposes segmentations independent of each other. While a pixel in the panoptic output can only be assigned to a single class and instance, instance segmentation proposals are often overlapping.

To handle these issues, Kirillov \etal \cite{kirillov2018panoptic} proposed a fusion process similar to non-maximum suppression (NMS) that favors instance proposals over semantic proposals. However, we observe that occlusion relationships between different objects do not correlate well with object detection confidences used in this NMS-like fusion procedure \cite{kirillov2018panoptic}, which therefore generally leads to poor performance when an instance that overlaps another (\eg, a tie on a shirt in Figure \ref{fig:COCO_occlusion}) has lower detection confidence than the instance it should occlude. This can cause a large number of instances that Mask R-CNN \textit{successfully} proposes fail to exist in the panoptic prediction (shown in Figure \ref{fig:fusion_process}). 

Therefore, in this work, we focus on enriching the fusion process established by \cite{kirillov2018panoptic} with a binary relationship between \textit{instances} to determine occlusion ordering. We propose adding an additional branch (occlusion head) to the instance segmentation pipeline tasked with determining which of two instance masks should lie on top of (or below) the other to resolve occlusions in the fusion process. The proposed occlusion head can be fine-tuned easily on top of an existing Panoptic Feature Pyramid Networks (FPNs) \cite{kirillov2019panoptic} architecture with minimal difficulty. We call our approach fusion with occlusion head (OCFusion). OCFusion brings significant performance gains on the COCO and Cityscapes panoptic segmentation benchmarks with low computational cost. 

\section{Learning Instance Occlusion for Panoptic Fusion}

We adopt the coupled approach of \cite{kirillov2019panoptic} that uses a shared Feature Pyramid Network (FPN) \cite{FPN} backbone with a top-down process for semantic segmentation branch and Mask R-CNN \cite{he2017mask} for instance segmentation branch.

In this section, we first discuss the instance occlusion problem arising within the fusion heuristic introduced in \cite{kirillov2018panoptic} and then introduce OCFusion method to address the problem. The overall approach is shown in Figure \ref{fig:overall_architecture}.

\subsection{Fusion by confidence} \label{fusing-instances}

The fusion protocol in \cite{kirillov2018panoptic} adopts a greedy strategy during inference in an iterative manner. Instance proposals are first sorted in order of decreasing detection confidence. In each iteration, the proposal is skipped if its intersection with the mask of all already assigned pixels is above a certain ratio of $\tau$. Otherwise, pixels in this mask that have yet to be assigned are assigned to the instance in the output. After all instance proposals of some minimum detection threshold are considered, the semantic segmentation is merged into the output by considering its pixels corresponding to each ``stuff'' class. If the number of pixels exceeds some threshold after removing already assigned pixels, then these pixels are assigned to the corresponding ``stuff" category. Pixels that are unassigned after this entire process are considered void predictions and have special treatment in the panoptic scoring process. We denote this type of fusion as \textit{fusion by confidence}.

\newpage

\begin{figure}[!htp]
\vspace{-1mm}
\begin{center}
\begin{subfigure}{0.47\textwidth}
\centering
\includegraphics[width=1.0\textwidth]{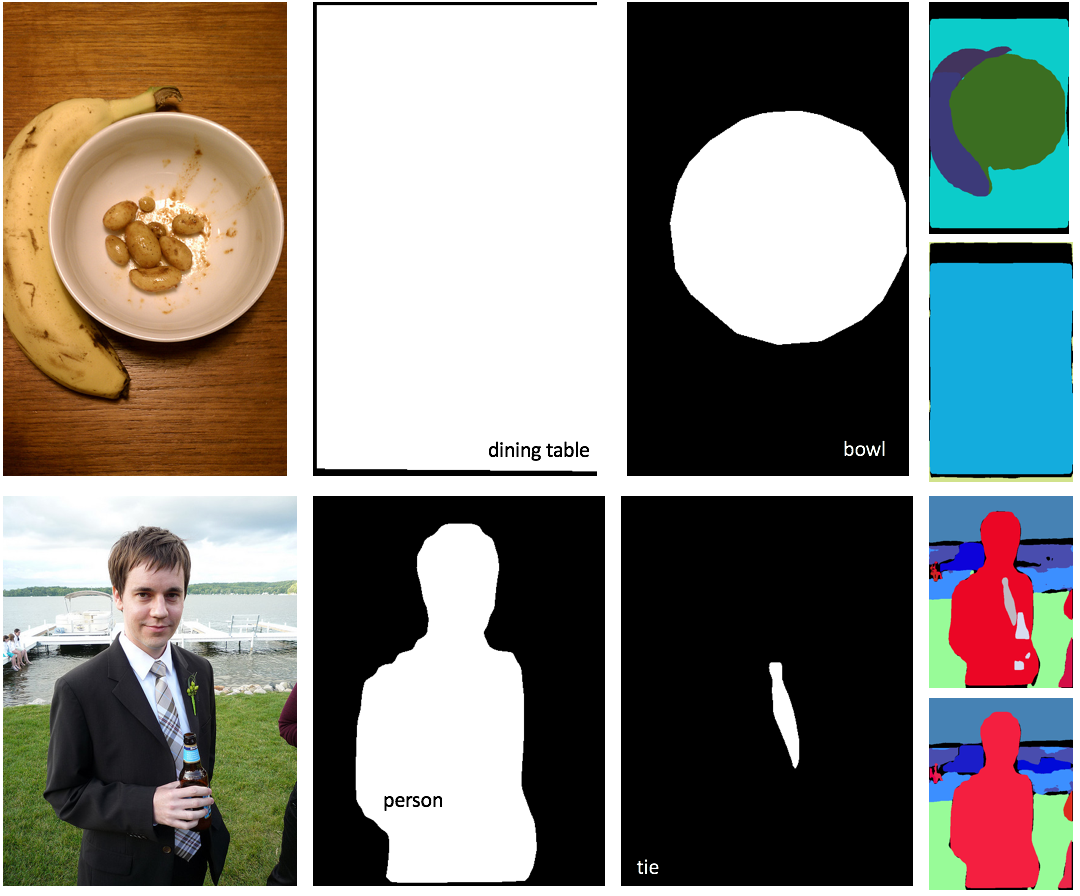} 
\caption{\phantom{a}}
\label{fig:COCO_occlusion}
\end{subfigure} \\
\begin{subfigure}{0.47\textwidth}
\centering
\includegraphics[width=1.0\textwidth]{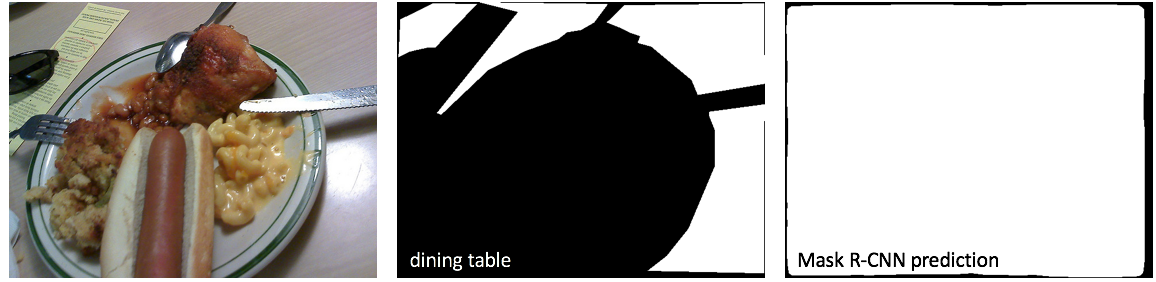}
\caption{\phantom{b}}
\label{fig:COCO_bad_prediction}
\end{subfigure}
\vspace{-3mm}
\caption{\textbf{Images and ground truth masks from the COCO dataset.} \textbf{(a)} is an example where even predicting the ground truth mask creates ambiguity when attempting to assign pixels to instances in a greedy manner. The \textbf{baseline fusion process} \cite{kirillov2018panoptic} is unable to properly assign these as shown in the \textbf{2nd and 4th} images of the rightmost column whereas \textbf{our method} is able to handle the occlusion relationship present as shown in the \textbf{1st and 3rd} images of the rightmost column. \textbf{(b)} is an example where Mask R-CNN baseline produces an instance prediction that occludes the entire image and creates the same ambiguity in (a) despite an unambiguous ground truth annotation.}
\end{center}
\vspace{-4mm}
\end{figure}

\noindent \textbf{Softening the greed.} The main weakness of the greedy fusion process is the complete reliance on detection confidences (\eg for Mask R-CNN, those from the box classification score) for a tangential task. Detection scores not only have little to do with mask quality (\eg, \cite{huang2019mask}), but they also do not incorporate any knowledge of \textit{layout}. If they are used in such a way, higher detection scores would imply a more foreground ordering. Often this is detrimental since Mask R-CNN exhibits behavior that can assign near-maximum confidence to very large objects (\eg see dining table images in Figure \ref{fig:COCO_bad_prediction}) that are both of poor mask quality and not truly foreground. It is common to see images with a significant number of true instances suppressed in the panoptic output by a single instance with large area that was assigned the largest confidence.

Our approach softens this greedy fusion process with an occlusion head that is dedicated to predicting the binary relation between instances with appreciable overlap so that instance occlusions can be properly handled.

\subsection{Occlusion head formulation}

Consider two masks $M_i$ and $M_j$ proposed by an instance segmentation model, and denote their intersection as $I_{ij} = M_i \cap M_j$. We are interested in the case where one of the masks is heavily occluded by the other. Therefore, we consider their respective intersection ratios $R_i = \text{Area}(I_{ij}) / \text{Area}(M_i)$ and $R_j = \text{Area}(I_{ij}) / \text{Area}(M_j)$ where $\text{Area}(M)$ denotes the number of ``on" pixels in mask $M$. As noted in Section \ref{fusing-instances}, the fusion process considers the intersection of the current instance proposal with the mask consisting of all already claimed pixels. Here, we are looking at the intersection between two masks and denote the threshold as $\rho$. If either $R_i \geq \rho$ or $R_j \geq \rho$, we define these two masks as having appreciable overlap. In this case,  we must then decide which instance the pixels in $I_{ij}$ should belong to. We attempt to answer this by learning a binary relation $\text{Occlude}(M_i, M_j)$  such that whenever $M_i$ and $M_j$ have appreciable intersection:

\vspace{-5mm}
\begin{align}
\label{occlude_equation}
\small
\text{Occlude}(M_i, M_j) =
\begin{cases}
    1 \text{ if } M_i \text{ should be placed on top of } M_j \\
    0 \text{ if } M_j \text{ should be placed on top of } M_i.
\end{cases}
\end{align}
\vspace{-5mm}

\subsection{Fusion with occlusion head}
\label{fusion-with_occlusion}
We now describe our modifications to the inference-time fusion heuristic of \cite{kirillov2018panoptic} that incorporates $\text{Occlude}(M_i, M_j)$ in Algorithm \ref{al:fusion-procedure}.

\begin{algorithm}[!htp]
\caption{\small Fusion with Occlusion Head.}
$P$ is $H \times W$ matrix, initially empty.\\
$\rho$ is a hyperparameter, the minimum intersection ratio for occlusion. \\
$\tau$ is a hyperparameter.

\label{al:fusion-procedure}
\begin{algorithmic}
{
{\small
\For {each proposed instance mask $M_i$}
    \State $C_i = M_i - P$ \Comment{pixels in $M_i$ that are not assigned in $P$}
    \For {$j < i$} \Comment{each already merged segment}
        \State{$I_{ij}$ is the intersection between mask $M_i$ and $M_j$}.
        \State{$R_{i} = \text{Area}(I_{ij}) / \text{Area}(M_i)$}.
        \State{$R_{j} = \text{Area}(I_{ij}) / \text{Area}(M_j)$}.        
        \If{$R_{i} \geq \rho$ or $R_j \geq \rho$} \Comment{significant intersection}
        \If{$\text{Occlude}(M_i, M_j) = 1$}
        \State{$C_i = C_i \bigcup \left( C_j \bigcap I_{ij}\right)$}.
        \State{$C_j = C_j - I_{ij}$}.
        \EndIf
        \EndIf
    \EndFor
    \If{$\text{Area}(C_i) / \text{Area}(M_i) \leq \tau$}
        \State{\textbf{continue}} 
    \Else
        \State{assign the pixels in $C_i$ to the panoptic mask} $P$.
    \EndIf
\EndFor
}
}
\end{algorithmic}

\end{algorithm}

After the instance fusion component has completed, the semantic segmentation is then incorporated as usual, only considering pixels assigned to \text{stuff} classes and determining whether the number of unassigned pixels corresponding to the class in the current panoptic output exceeds some threshold, \eg, 4096. The instance fusion process is illustrated in Figure \ref{fig:fusion_process}.

\subsection{Occlusion head architecture}
\label{sec:architecture_of_occlusion_head}
We implement $\text{Occlude}(M_i, M_j)$ as an additional ``head'' within Mask R-CNN \cite{he2017mask}. Mask R-CNN already contains two heads: a box head that is tasked with taking region proposal network (RPN) proposals and refining the bounding box as well as assigning classification scores, while the mask head predicts a fixed size binary mask (usually $28 \times 28$) for all classes independently from the output of the box head. Each head derives its own set of features from the underlying FPN. We name our additional head, the ``occlusion head" and implement it as a binary classifier that takes two (soft) masks $M_i$ and $M_j$ along with their respective FPN features (determined by their respective boxes) as input. The classifier output is interpreted as the value of $\text{Occlude}(M_i, M_j)$.

The architecture of occlusion head is inspired by \cite{huang2019mask} as shown in Figure \ref{fig:overall_architecture}. For two mask representations $M_i$ and $M_j$, we apply max pooling to produce a $14 \times 14$ representation and concatenate each with the corresponding RoI features to produce the input to the head. Three layers of $3 \times 3$ convolutions with 512 feature maps and stride 1 are applied before a final one with stride 2. The features are then flattened before a 1024 dimensional fully connected layer and finally projected to a single logit.

\subsection{Ground truth occlusion} \label{sec:gt_occlusion}

We use ground truth panoptic mask along with ground truth instance masks to derive ground truth occlusion relation. We pre-compute the intersection between all pairs of masks with appreciable overlap. We then find the pixels corresponding to the intersection of the masks in the panoptic ground truth. We determine the instance occlusion based on which instance owns the majority of pixels in the intersection. We store the resulting ``occlusion matrix" for each image in an $N_i \times N_i$ matrix where $N_i$ is the number of instances in the image and the value at position $(i, j)$ is either $-1$, indicating no occlusion, or encodes the value of $\text{Occlude}(i, j)$.

\subsection{Occlusion head training}

During training, the occlusion head is designed to first find pairs of predicted masks that match to different ground truth instances. Then, the intersection between these pairs of masks is computed, and the ratio of the intersection to mask area taken. A pair of masks is added for consideration when one of these ratios is at least as large as the pre-determined threshold $\rho$. We then subsample the set of all pairs meeting this criterion to decrease computational cost. It is desirable for the occlusion head to reflect the consistency of $\text{Occlude}$, therefore we also invert all pairs so that $\text{Occlude}(M_i, M_j) = 0 \iff \text{Occlude}(M_j, M_i) = 1$ whenever the pair $(M_i, M_j)$ meets the intersection criteria. This also mitigates class imbalance. Since this is a binary classification problem, the overall loss $L_o$ from the occlusion head is given by the binary cross-entropy over all subsampled pairs of masks that meet the intersection criteria.

\section{Related work}
\label{sec:related_work}

Next, we discuss in detail the difference between OCFusion and the existing approaches for panoptic segmentation, occlusion ordering, and non-maximum suppression.\\

\noindent \textbf{Panoptic segmentation}. The task of panoptic segmentation is introduced in \cite{kirillov2018panoptic} along with a baseline where predictions from instance segmentation (Mask R-CNN \cite{he2017mask}) and semantic segmentation (PSPNet \cite{zhao2016pspnet}) are combined via a heuristics-based fusion strategy. A stronger baseline based on a single Feature Pyramid Network (FPN) \cite{FPN} backbone followed by multi-task heads consisting of semantic and instance segmentation branches is concurrently proposed by \cite{li2018attention, li2018learning, kirillov2019panoptic, xiong2019upsnet}. On top of this baseline,  attention layers are added in \cite{li2018attention} to the instance segmentation branch, which are guided by the semantic segmentation branch;  a loss term enforcing consistency between things and stuff predictions is then introduced in \cite{li2018learning};  a parameter-free panoptic head which computes the final panoptic mask by pasting instance mask logits onto semantic segmentation logits is presetned in \cite{xiong2019upsnet}. These works have been making steady progress in panoptic segmentation, but their focus is not to address the problem for explicit reasoning of instance occlusion.

\noindent \textbf{Occlusion ordering and layout learning.}
Occlusion handling is a long-studied computer vision task \cite{wang2009hog,enzweiler2010multi,sun2005symmetric,hoiem2007recovering}. In the context of semantic segmentation, occlusion ordering has been adopted in \cite{tighe2014scene,Chen_2015_CVPR,zhu2017semantic}. A repulsion loss is added to a pedestrian detection algorithm \cite{wang2018repulsion} to deal with the crowd occlusion problem, but it focuses on detection only, without instance segmentation.
\noindent In contrast, we study the occlusion ordering problem for instance maps in panoptic segmentation. Closest to our method is the recent work of \cite{liu2019end}, which proposes a panoptic head to resolve this issue in a similar manner to \cite{xiong2019upsnet} but instead with a learnable convolution layer. Since our occlusion head can deal with two arbitrary masks, it offers more flexibility over these approaches which attempt to ``rerank" the masks in a linear fashion \cite{xiong2019upsnet, liu2019end}. Furthermore, the approach of \cite{liu2019end} is based off how a \textit{class} should be placed on top of \textit{another class} (akin to semantic segmentation) while we explicitly model the occlusion relation between arbitrary \textit{instances}. This allows us to leverage the \textit{intra-class occlusion relations} such as ``which of these two persons should occlude the other?'', and we show this leads to a gain in Figure \ref{fig:intra-class-cityscapes} and Table \ref{table:type_occlusion_ablation}. In a nutshell, we tackle the occlusion problem in a scope that
is more general than \cite{liu2019end} with noticeable performance advantage, as shown in Table \ref{table:coco_comp_val} and Table \ref{table:testdev}. 

\noindent\textbf{Learnable NMS}. One can relate resolving occlusions to non-maximum suppression (NMS) that is applied to \textit{boxes}, while our method tries to suppress intersections between masks. Our method acts as a \textit{learnable} version of NMS for instance masks with similar computations to the analogous ideas for boxes such as \cite{hosang2017learning}.

\section{Experiments}
\subsection{Implementation details}
We extend the Mask R-CNN benchmark framework \cite{massa2018mrcnn}, built on top of PyTorch, to implement our architecture. Batch normalization \cite{ioffe2015batch} layers are frozen and not fine-tuned for simplicity. We perform experiments on the COCO dataset \cite{lin2014microsoft}  \cite{kirillov2018panoptic} as well as the Cityscapes dataset \cite{Cordts2016Cityscapes} with panoptic annotations. 

We find the most stable and efficient way to train the occlusion head is by fine-tuning with all other parameters frozen. We add a single additional loss only at fine-tuning time so that the total loss during panoptic training is $L = \lambda_i(L_c + L_b + L_m) + \lambda_s L_s$ where $L_c$, $L_b$, and $L_m$ are the box head classification loss, bounding-box regression loss, and mask loss while $L_s$ is the semantic segmentation cross-entropy loss. At fine-tuning time, we only minimize $L_o$, the classification loss from the occlusion head. We subsample 128 mask occlusions per image.

During fusion, we only consider instance masks with detection confidence of at least $0.5$ or $0.6$ and reject segments during fusion when their overlap ratio with the existing panoptic mask (after occlusions are resolved) exceeds $\tau = 0.5$ on COCO and $\tau = 0.6$ on Cityscapes. Lastly, when considering the segments of \textit{stuff} generated from the semantic segmentation, we only consider those which have at least 4096 pixels remaining after discarding those already assigned on COCO and 2048 on Cityscapes.\\
\noindent \textbf{Semantic head.} On COCO, repeat the combination of $3\times3$ convolution and $2\times$ bilinear upsampling until $1/4$ scale is reached, following the design of \cite{kirillov2019panoptic}. For the model with ResNeXt-101 backbone, we replace each convolution with deformable convolution \cite{deformable-conv}. For ResNet-50 backbone, we additionally add one experiment that adopts the design from \cite{xiong2019upsnet} which uses 2 layers of deformable convolution followed by a bilinear upsampling to the $1/4$ scale. On Cityscapes, we adopt the design from \cite{xiong2019upsnet}. \\
\noindent \textbf{COCO.} The COCO 2018 panoptic segmentation task consists of 80 \textit{thing} and 53 \textit{stuff} classes. We use 2017 dataset which has a split of 118k/5k/20k for training, validation and testing respectively.\\
\noindent \textbf{Cityscapes.} Cityscapes consists of 8 \textit{thing} classes and 11 \textit{stuff} classes. We use only \textit{fine} dataset with a split of 2975/500/1525 for training, validation and testing respectively.\\
\noindent \textbf{COCO training.} We train the FPN-based architecture described in \cite{kirillov2019panoptic} for 90K iterations on 8 GPUs with 1 image per GPU. The base learning rate of 0.02 is reduced by 10 at both 60k and 80k iterations. We then proceed to fine-tune with the occlusion head for 2500 more iterations. We choose $\lambda_i = 1.0$ and $\lambda_s = 0.5$ while for the occlusion head we choose the intersection ratio $\rho$ as 0.2. For models with ResNet-50 and ResNet-101 backbone, we use random horizontal flipping as data augmentation. For model with ResNeXt-101 backbone, we additionally use scale jitter (with scale of shorter image edge equals to $\{640, 680, 720, 760, 800\}$).\\
\noindent \textbf{Cityscapes training.} We randomly rescale each image by 0.5 to 2$\times$ (scale factor sampled from a uniform distribution) and construct each batch of 16 (4 images per GPU) by randomly cropping images of size 512 $\times$ 1024. We train for 65k iterations with a base learning rate of 0.01 with decay at 40k and 55k iterations. We fine-tune the occlusion head for 5000 more iterations. We choose $\lambda_i = \lambda_s = 1.0$ with intersection ratio $\rho$ as 0.1. We do not pretrain on COCO.\\ 
\noindent \textbf{Panoptic segmentation metrics.} We adopt the panoptic quality (PQ) metric from \cite{kirillov2018panoptic} to measure panoptic segmentation performance. This single metric captures both segmentation and recognition quality. PQ can be further broken down into scores specific to \textit{things} and \textit{stuff}, denoted PQ$\textsuperscript{Th}$ and PQ$\textsuperscript{St}$, respectively.\\
\noindent \textbf{Multi-scale testing.} We adopt the same scales as \cite{xiong2019upsnet} for both COCO and Cityscapes multi-scale testing. For the stuff branch, we average the multi-scale semantic logits of semantic head. For the thing branch, we average the multi-scale masks and choose not to do bounding box augmentation for simplicity.

\begin{table}[!htp]
\vspace{-3mm}
\centering
\setlength{\tabcolsep}{3.0pt}
\begin{tabular}{@{}lccccc@{}}
\hline
\toprule
\textbf{Method} & \textbf{Backbone} & \textbf{PQ} & \textbf{PQ}\textsuperscript{Th} & \textbf{PQ}$\textsuperscript{St}$ \\
\midrule
Baseline & ResNet-50 & 39.5 & 46.5 & 29.0 \\
OCFusion & ResNet-50 & 41.3 & 49.4 & 29.0 \\
\midrule
relative improvement & {} & +1.8 & +3.0 & +0.0 \\
\midrule
Baseline & ResNet-101 & 41.0 & 47.9 & 30.7 \\
OCFusion & ResNet-101 & 43.0 & 51.1 & 30.7 \\
\midrule
relative improvement & {} & +2.0 & +3.2 & +0.0 \\

\bottomrule
\hline
\end{tabular}
\vspace{-2mm}
\caption{\textbf{Comparison to our implementation of Panoptic FPN \cite{kirillov2019panoptic} baseline model on the MS-COCO \textit{val} dataset}.}
\label{table:coco_comp_baseline}
\vspace{-0mm}
\end{table}

\begin{table}[tb]
\centering

\setlength{\tabcolsep}{2.5pt}
\begin{tabular}{@{}lccccc@{}}

\hline
\toprule
&&&&\\[-1em]
\textbf{Method} & \textbf{Backbone} & \textbf{\thead{m.s. \\ test}} & \textbf{PQ} &\textbf{PQ}\textsuperscript{Th} & \textbf{PQ}$\textsuperscript{St}$ \\
\midrule
JSIS-Net \cite{de2018panoptic} & ResNet-50 & {} & 26.9 & 29.3 & 23.3 \\
Panoptic FPN \cite{kirillov2019panoptic} & ResNet-50 & {} & 39.0 & 45.9 & 28.7 \\
Panoptic FPN \cite{kirillov2019panoptic} & ResNet-101 & {} & 40.3 & 47.5 & 29.5 \\
AUNet \cite{li2018attention} & ResNet-50 & {} & 39.6 & 49.1 & 25.2 \\
UPSNet\textsuperscript{$\ast$} \cite{xiong2019upsnet} & ResNet-50 & {} & 42.5 & 48.5 & 33.4 \\
UPSNet\textsuperscript{$\ast$} \cite{xiong2019upsnet} & ResNet-50 & {\checkmark} & 43.2 & 49.1 & 34.1 \\
OANet \cite{liu2019end} & ResNet-50 & {} & 39.0 & 48.3 & 24.9 \\
OANet \cite{liu2019end} & ResNet-101 & {} & 40.7 & 50.0 & 26.6 \\
AdaptIS \cite{sofiiuk2019adaptis} & ResNet-50 & {} & 35.9 & 40.3 & 29.3 \\
AdaptIS \cite{sofiiuk2019adaptis} & ResNet-101 & {} & 37 & 41.8 & 29.9 \\
AdaptIS \cite{sofiiuk2019adaptis} & ResNeXt-101 & {} & 42.3 & 49.2 & 31.8 \\
\midrule
OCFusion & ResNet-50 & {} & 41.3 & 49.4 & 29.0 \\
OCFusion\textsuperscript{$\ast$} & ResNet-50 & {} & 42.5 & 49.1 & 32.5 \\
OCFusion & ResNet-101 & {} & 43.0 & 51.1 & 30.7 \\

OCFusion\textsuperscript{$\ast$} & ResNeXt-101 & {} & 45.7 & 53.1 & 34.5 \\

OCFusion & ResNet-50 & \checkmark & 41.9 & 49.9 & 29.9 \\
OCFusion\textsuperscript{$\ast$} & ResNet-50 & \checkmark & 43.3 & 50.0 & 33.8 \\
OCFusion & ResNet-101 & \checkmark & 43.5 & 51.5 & 31.5 \\
OCFusion\textsuperscript{$\ast$} & ResNeXt-101 & \checkmark & \textbf{46.3} & \textbf{53.5} & \textbf{35.4} \\
\bottomrule
\hline
\end{tabular}
\vspace{-2mm}
\caption{\textbf{Comparison to prior work on the MS-COCO \textit{val} dataset.} m.s. stands for multi-scale testing. \textsuperscript{$\ast$}Used deformable convolution. }
\label{table:coco_comp_val}
\vspace{-1.mm}
\end{table}

\begin{table}[!htp]
\centering
\setlength{\tabcolsep}{2.5pt}
\begin{tabular}{@{}lccccc@{}}
\hline
\toprule
\textbf{Method} & \textbf{Backbone} & \textbf{\thead{m.s. \\ test}} & \textbf{PQ}  & \textbf{PQ}\textsuperscript{Th} & \textbf{PQ}$\textsuperscript{St}$ \\
\midrule

JSIS-Net \cite{de2018panoptic} & ResNet-50 & {} & 27.2 & 29.6 & 23.4 \\
Panoptic FPN \cite{kirillov2019panoptic} & ResNet-101 & {} & 40.9 & 48.3 & 29.7 \\
OANet \cite{liu2019end} & ResNet-101 & {} & 41.3 & 50.4 & 27.7 \\
AUNet \cite{li2018attention} & ResNeXt-152 & \checkmark & 46.5 & \textbf{55.9} & 32.5 \\
UPSNet\textsuperscript{$\ast$} \cite{xiong2019upsnet} & ResNet-101 & \checkmark&  46.6 & 53.2 & 36.7 \\
AdaptIS \cite{sofiiuk2019adaptis} & ResNeXt-101 & {} & 42.8 & 50.1 & 31.8 \\
\midrule
OCFusion\textsuperscript{$\ast$} & ResNeXt-101 & \checkmark  & \textbf{46.7} & 54.0 & \textbf{35.7} \\

\bottomrule
\hline
\end{tabular}
\vspace{-2mm}
\caption{\textbf{Comparison to prior work on the MS-COCO \textit{test-dev} dataset.} m.s. stands for multi-scale testing. \textsuperscript{$\ast$}Used deformable convolution.}
\label{table:testdev}
\vspace{0mm}
\end{table}

\subsection{COCO panoptic benchmark}

We obtain state-of-the-art results on COCO Panoptic Segmentation validation set with and without multi-scale testing as is shown in \ref{table:coco_comp_val}. We also obtain single model state-of-the-art results on the COCO test-dev set, as shown in Table \ref{table:testdev}. In order to show the effectiveness of our method, we compare to our baseline model in Table \ref{table:coco_comp_baseline}, and the results show that our method consistently provides significant gain on PQ\textsuperscript{Th} as well as PQ.

\subsection{Cityscapes panoptic benchmark}
We obtain competitive results on the Cityscapes validation set and the best results among models with a ResNet-50 backbone, shown in Table \ref{table:cityscapes_comp_val}. Table \ref{table:cityscapes_comp_baseline} shows our strong relative improvement over the baseline on PQ\textsuperscript{Th} as well as PQ. 
\begin{table}[!htp]
\centering
\vspace{0.05cm}
\vspace{-2mm}
\scalebox{1.00}{
\setlength{\tabcolsep}{5.0pt}
\begin{tabular}{@{}lccc@{}}
\hline
\toprule
\textbf{Method} & \textbf{PQ} & \textbf{PQ}\textsuperscript{Th} & \textbf{PQ}$\textsuperscript{St}$ \\
\midrule
Baseline & 58.6 & 51.7 & 63.6 \\
OCFusion & 59.3 & 53.5 & 63.6 \\
\midrule
 relative improvement & +0.7 & +1.7 & +0.0 \\
\bottomrule
\hline
\end{tabular}
}
\vspace{-2mm}
\caption{\textbf{Comparison to our implementation of Panoptic FPN \cite{kirillov2019panoptic} baseline model on the Cityscapes \textit{val} dataset}. All results are based on a ResNet-50 backbone.}
\label{table:cityscapes_comp_baseline}
\vspace{-0mm}
\end{table}

\begin{table}[!htp]
\vspace{-0mm}
\centering
\vspace{-3mm}
\setlength{\tabcolsep}{5.0pt}
\begin{tabular}{@{}lcccc@{}}
\hline
\toprule
\textbf{Method} & \textbf{\thead{m.s. \\ test}} & \textbf{PQ} &\textbf{PQ}\textsuperscript{Th} & \textbf{PQ}$\textsuperscript{St}$ \\
\midrule
Panoptic FPN \cite{kirillov2019panoptic} & {} & 57.7 & 51.6 & 62.2 \\
AUNet \cite{li2018attention} & {} & 56.4 & 52.7 & 59.0 \\
UPSNet\textsuperscript{$\ast$} \cite{xiong2019upsnet} & {} & 59.3 & 54.6 & 62.7 \\
UPSNet\textsuperscript{$\ast$} \cite{xiong2019upsnet} & {\checkmark} & 60.1 & 55.0 & 63.7 \\
AdaptIS \cite{sofiiuk2019adaptis} & {} & 59.0 & \textbf{55.8} & 61.3 \\

\midrule
OCFusion\textsuperscript{$\ast$} & {} & 59.3 & 53.5 & 63.6 \\
OCFusion\textsuperscript{$\ast$} & \checkmark & \textbf{60.2} & 54.0 & \textbf{64.7} \\
\bottomrule
\hline
\end{tabular}

\vspace{-2mm}
\caption{\textbf{Comparison to prior work on the Cityscapes \textit{val} dataset.} All results are based on a ResNet-50 backbone. m.s. stands for multi-scale testing. \textsuperscript{$\ast$}Used deformable convolution. }
\label{table:cityscapes_comp_val}
\vspace{-2mm}
\end{table}

\begin{figure*}

\vspace{0mm}
 \begin{center}
\begin{tabular} {c}
\hspace{0mm} \includegraphics[width=0.75\textwidth]{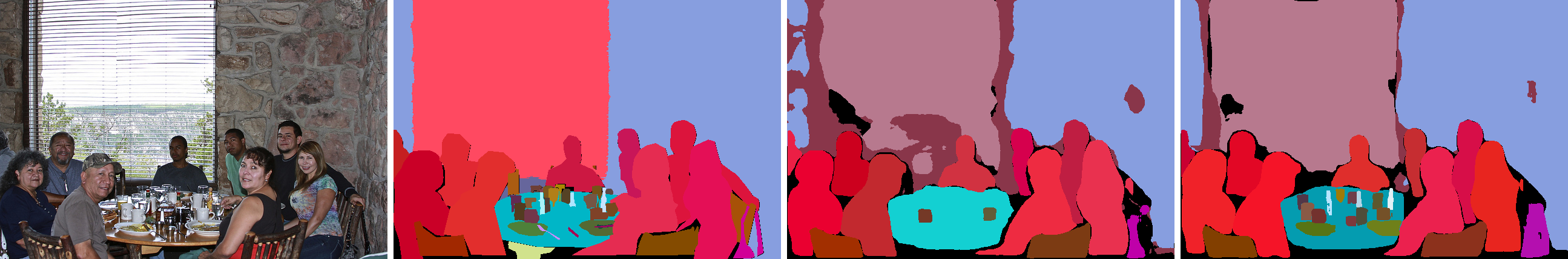} \\
\hspace{0mm} \includegraphics[width=0.75\textwidth]{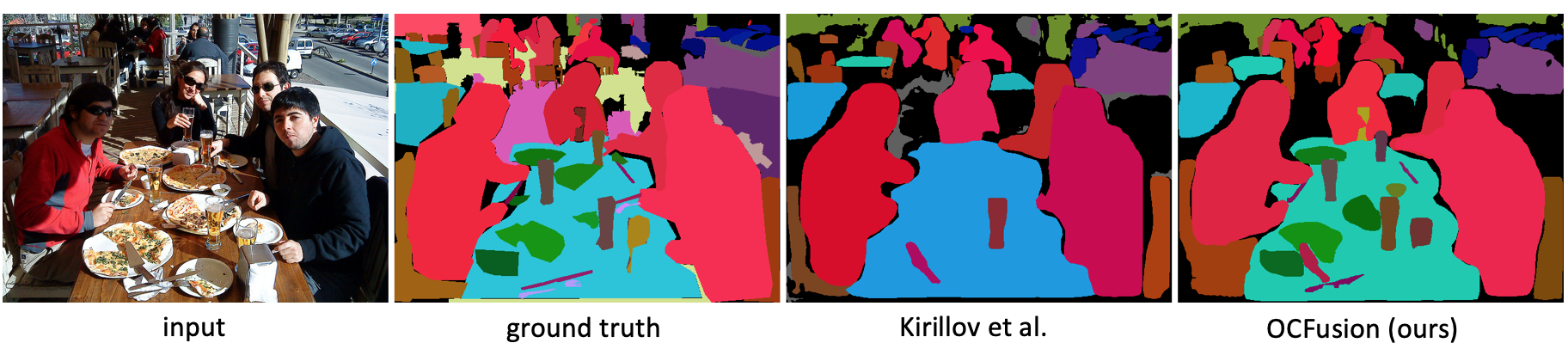}
\end{tabular}
 \end{center}
\vspace{-6mm} 
\caption{\textbf{Comparison against Kirillov et al. \cite{kirillov2019panoptic} which uses fusion by confidence.}}
\vspace{5mm}
\label{fig:figure_ComparisonBaseline}
\end{figure*}

\begin{figure*}[!htp]
\vspace{-0mm}
\begin{center}
\begin{tabular} {c}
\hspace{-5mm} \includegraphics[width=0.95\textwidth]{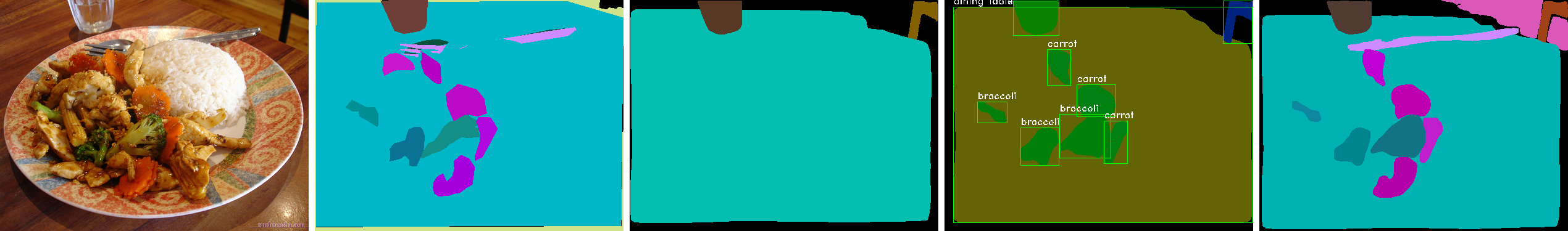} \\
\hspace{-5mm} \includegraphics[width=0.95\textwidth]{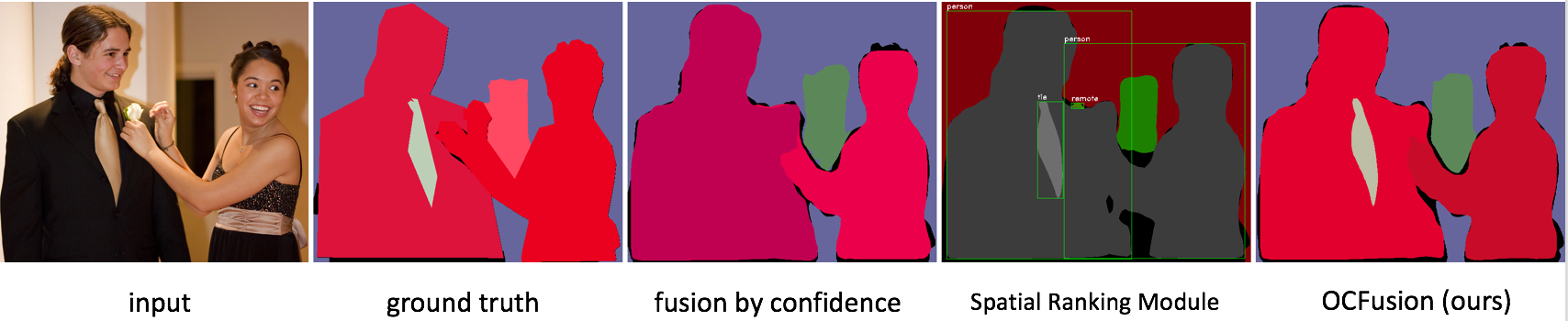}
\end{tabular}
\end{center}
\vspace{-6mm}
\caption{\textbf{Comparison against Spatial Ranking Module \cite{liu2019end}.}}
\vspace{5mm}
\label{fig:figure_ComparisonEndToEnd}
\end{figure*}

\begin{figure*}[!h]

\vspace{0mm}
\begin{center}
\begin{tabular} {c}
\hspace{-5mm} \includegraphics[width=0.95\textwidth]{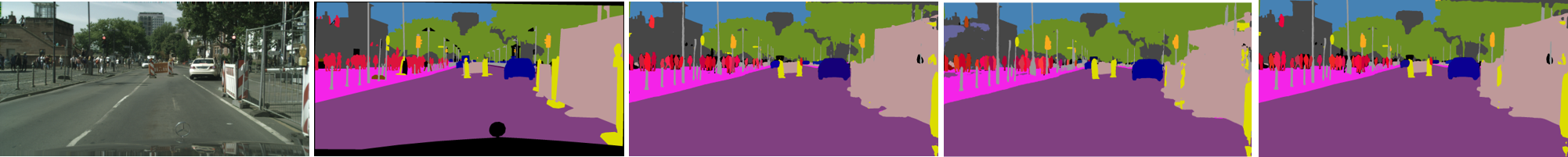} \\
\hspace{-5mm} \includegraphics[width=0.95\textwidth]{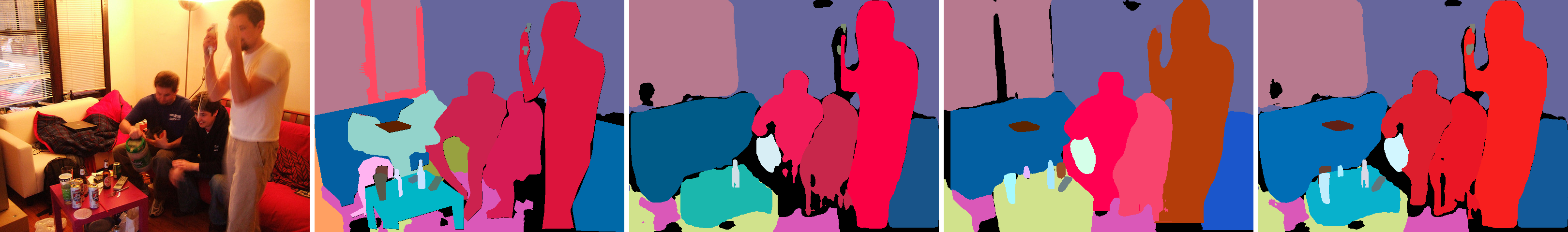} \\
\hspace{-5mm} \includegraphics[width=0.95\textwidth]{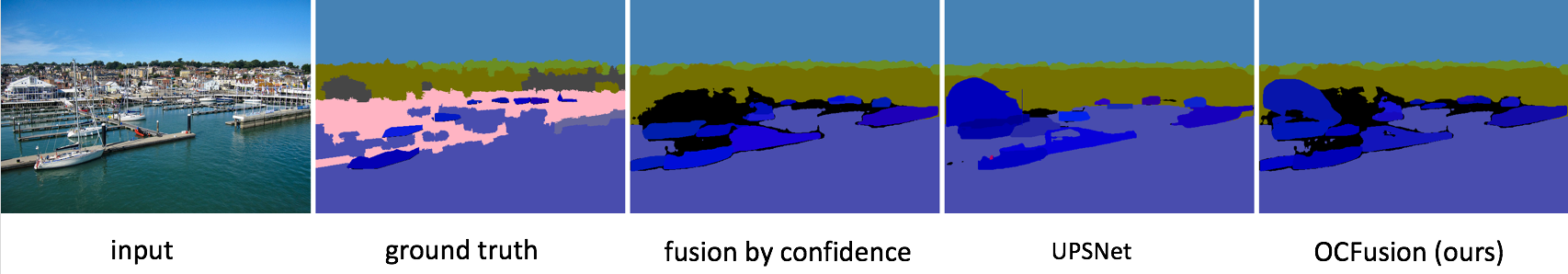}
\end{tabular}
\end{center}
\vspace{-6mm}
\caption{\textbf{Comparison against UPSNet \cite{xiong2019upsnet}.}}
\vspace{0mm}
\label{fig:figure_ComparisonUPSNet}
\end{figure*}

\subsection{Occlusion head performance}
In order to better gauge the performance of the occlusion head, we determine its classification accuracy on both COCO and Cityscapes validation dataset at $\rho = 0.20$ with ResNet-50 backbone. We measure the accuracy of the occlusion head in predicting the true ordering given ground truth boxes and masks. The occlusion head classification accuracy on COCO and Cityscapes is 91.58\% and 93.60\%, respectively, which validates the effectiveness of OCFusion.

\subsection{Inference time analysis}
We analyze the computational cost of our method and empirically show the inference time overhead of our method compared to the baseline model. While our method incurs an $O(n^2)$ cost in order to compute pairwise intersections, where $n$ is the number of instances, this computation is only needed for the subset of masks whose detection confidence is larger than a threshold (0.5 or 0.6 usually) as dictated by the Panoptic FPN \cite{kirillov2019panoptic} baseline. This filtering greatly limits the practical magnitude of $n$. Furthermore, only the subset of remaining mask pairs that have appreciable overlap (larger than $\rho$) requires evaluation by the occlusion head. We measure this inference time overhead in Table \ref{table:inference_time_analysis}. OCFusion incurs a modest 2.0\% increase in computational time on COCO and 4.7\% increase on Cityscapes.

\vspace{-1.5mm}
\begin{table}[!htp]
\vspace{-0mm}
\centering
\scalebox{1.0}{
\setlength{\tabcolsep}{5.0pt}
\begin{tabular}{@{}lcc@{}}
\hline
\toprule
\textbf{Method} & \textbf{COCO} & \textbf{Cityscapes} \\
\midrule
Baseline & 153 & 378  \\
OCFusion & 156 & 396 \\
Change in runtime (ms) & +3 & +18 \\
\bottomrule
\hline
\end{tabular}
}
\vspace{-2mm}
\caption{ \textbf{Runtime (ms) overhead per image.} Runtime results are averaged over the entire COCO and Cityscapes validation dataset. We use a single GeForce GTX 1080 Ti GPU and Xeon(R) CPU E5-2687W CPU.
} 
\label{table:inference_time_analysis}
\vspace{-5mm}
\end{table}

\subsection{Visual comparisons}

Since panoptic segmentation is a relatively new task, the most recent papers offer only comparisons against the baseline presented in \cite{kirillov2018panoptic}. We additionally compare with a few other recent methods \cite{liu2019end,xiong2019upsnet}.

We first compare our method against \cite{kirillov2019panoptic} in Figure \ref{fig:figure_ComparisonBaseline} as well as two recent works: UPSNet \cite{xiong2019upsnet} in Figure \ref{fig:figure_ComparisonUPSNet} and the Spatial Ranking Module of \cite{liu2019end} in Figure \ref{fig:figure_ComparisonEndToEnd}. The latter two have similar underlying architectures alongside modifications to their fusion process. We note that except for comparisons between \cite{kirillov2019panoptic}, the comparison images shown are those \textit{included in the respective papers and not of our own choosing}. Overall, we see that our method is able to preserve a significant number of instance occlusions lost by other methods while maintaining more realistic fusions, \eg, the arm is entirely above the man versus sinking behind partly as in ``fusion by confidence".

\vspace{1mm}

\begin{figure}[!htp]
\begin{center}
\begin{tabular} {c}
\includegraphics[width=0.45\textwidth]{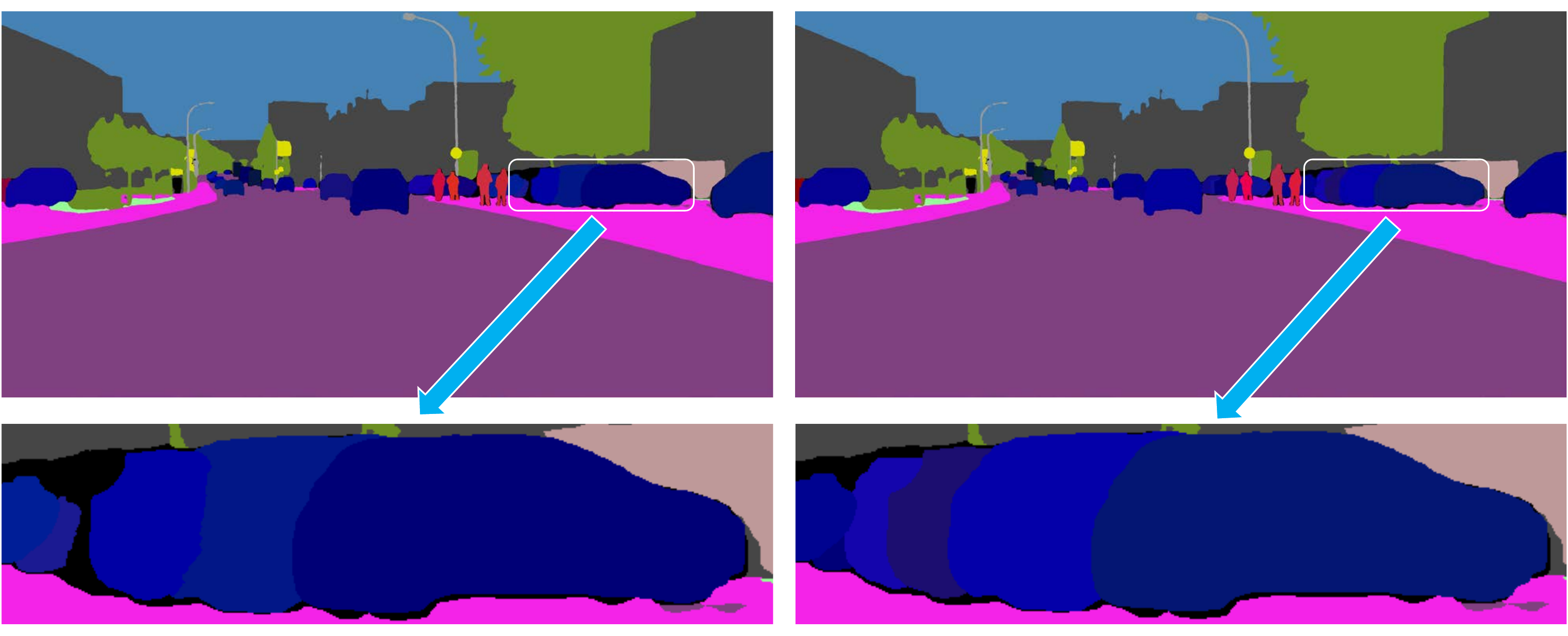}
\end{tabular}
\end{center}
\vspace{-7mm}
\caption{\textbf{Comparison for w/o (left) or w/ (right) intra-class capability enabled.} Best viewed in color.} 
\vspace{-3mm}
\label{fig:intra-class-cityscapes}
\end{figure}

\subsection{Ablation experiments} \label{sec:ablation}

We study the sensitivity of our method to the hyperparameters $\tau$ and $\rho$ in Table \ref{table:ablation_coco} for COCO and Table \ref{table:ablation_cityscapes} for Cityscapes. We also include the number of examples of occlusions we are able to collect at the given $\rho$ denoted as N. Naturally, a larger $\rho$ leads to less spurious occlusions but decreases the overall number of examples that the occlusion head is able to learn from.

Intra-class instance occlusion in Cityscapes is a challenging problem, also noted in \cite{he2017mask}. Since we can enable inter-class or intra-class occlusion query ability independently, we show ablation results in Table \ref{table:type_occlusion_ablation} that highlight the importance of being able to handle intra-class occlusion on. We believe this sets our method apart from others, \eg, \cite{liu2019end} which simplifies the problem by handling inter-class occlusion only. Additionally, Figure \ref{fig:intra-class-cityscapes} shows a visual comparison between resulting panoptic segmentations when intra-class occlusion handling is toggled on Cityscapes. Only the model with intra-class handling enabled can handle the occluded cars better during the fusion process.

\begin{table}
\centering
\scalebox{0.62}{
\setlength{\tabcolsep}{2.5pt}
\begin{tabular}{@{}llll@{}}
\hline
\toprule
 ($\tau$, $\rho$) & 0.05 & 0.10 & 0.20 \\
 \hline
 0.4 & 41.27 (Th: 49.43, St: 28.97) & 41.22 (Th: 49.33, St: 28.97) & 41.20 (Th: 49.30, St: 28.97) \\
 0.5 & 41.20 (Th: 49.32, St: 28.95) & 41.15 (Th: 49.23, St: 28.95) & 41.24 (Th: 49.29, St: 29.10) \\
 0.6 & 41.09 (Th: 49.15, St: 28.93) & 41.03 (Th: 49.03, St: 28.93) & 41.02 (Th: 49.02, St: 28.93) \\
 \hline
 N  & 192,519 & 157,784 & 132,165 \\
\bottomrule
\hline
\end{tabular}
}
\vspace{-2mm}
\caption{\footnotesize \textbf{COCO Hyperparameter Ablation: PQ}}

\label{table:ablation_coco}
\vspace{-3mm}
\end{table}

\begin{table}
\centering
\scalebox{0.62}{
\setlength{\tabcolsep}{2.5pt}
\begin{tabular}{@{}llll@{}}
\hline
\toprule
 ($\tau$, $\rho$) & 0.05 & 0.10 & 0.20 \\
 \hline
 0.4 & 58.76 (Th: 52.10, St: 63.62) & 59.15 (Th: 53.00, St: 63.62) & 59.07 (Th: 52.80, St: 63.63)\\
 0.5 & 59.18 (Th: 53.09, St: 63.61) & 59.26 (Th: 53.28, St: 63.61) & 59.22 (Th: 53.19, St: 63.61) \\
 0.6 & 59.21 (Th: 53.17, St: 63.61) & 59.33 (Th: 53.46, St: 63.60) & 58.70 (Th: 51.96, St: 61.60) \\
 \hline
 N & 33,391 & 29,560 & 6,617 \\
\bottomrule
\hline
\end{tabular}
}
\vspace{-2mm}
\caption{\footnotesize \textbf{Cityscapes Hyperparameter Ablation: PQ}}
\label{table:ablation_cityscapes}
\vspace{-4mm}
\end{table}

\begin{table}[!htp]
\centering

\resizebox{\linewidth}{!}{
\setlength{\tabcolsep}{2.3pt}
\begin{tabular}{@{}ccccc@{}}
\hline
\toprule
\textbf{Inter-class} & \textbf{Intra-class} & \textbf{PQ} &\textbf{PQ}\textsuperscript{Th} & \textbf{PQ}$\textsuperscript{St}$ \\
\midrule
{} & {} & 58.6 & 51.7 & 63.6 \\
\checkmark & {} & 59.2 (+0.5) & 53.0 (+1.3) & 63.6 (+0.0) \\
\checkmark & \checkmark & 59.3 (+0.7) & 53.5 (+1.7) & 63.6 (+0.0)\\
\bottomrule
\hline
\end{tabular}

}
\vspace{-2mm}
\caption{\textbf{Ablation study on different types of occlusion on the Cityscapes \textit{val} dataset.} \checkmark means capability enabled.}
\label{table:type_occlusion_ablation}
\vspace{0mm}
\end{table}

\vspace{-5mm}

\section{Conclusion}
We have introduced an \textit{explicit} notion of instance occlusion to Mask R-CNN so that instances may be effectively fused when producing a panoptic segmentation. We assemble a dataset of occlusions already present in the COCO and Cityscapes datasets and then learn an additional head for Mask R-CNN tasked with predicting occlusion between two masks. Adding occlusion head on top of Panoptic FPN incurs minimal overhead, and we show that it is effective even when trained for few thousand iterations. In the future, we hope to explore how further understanding of occlusion, including relationships of \textit{stuff}, could be helpful.\\

\noindent {\bf Acknowledgements}.
This work is supported by NSF IIS-1618477 and IIS-1717431. We thank Yifan Xu, Weijian Xu, Sainan Liu, Yu Shen, and Subarna Tripathi for valuable discussions. Also, we appreciate the anonymous reviewers for their helpful and constructive comments and suggestions.

{\small
\bibliographystyle{ieee_fullname}
\bibliography{bib}
}

\end{document}